\documentclass{article}

\usepackage{nips97e}

\usepackage[bf,hang]{subfigure}
\usepackage{graphicx}
\usepackage[bf,hang]{caption}

\title{USING COLLECTIVE INTELLIGENCE TO ROUTE INTERNET TRAFFIC}

\author{David H. Wolpert\\
NASA Ames Research Center\\
Moffett Field, CA 94035\\
dhw@ptolemy.arc.nasa.gov\\
\And
Kagan Tumer\\
NASA Ames Research Center\\
Caelum Research \\
Moffett Field, CA 94035\\
kagan@ptolemy.arc.nasa.gov\\
\AND
Jeremy Frank\\
NASA Ames Research Center\\
Caelum Research \\
Moffett Field, CA 94035\\
frank@ptolemy.arc.nasa.gov\\
}

\begin{document}

\vspace*{-.3in}

\maketitle

\begin{abstract}

A COllective INtelligence (COIN) is a set of interacting reinforcement
learning (RL) algorithms designed in an automated fashion so that
their collective behavior optimizes a global utility function. We
summarize the theory of COINs, then present experiments using that
theory to design COINs to control internet traffic routing. These
experiments indicate that COINs outperform all previously investigated
RL-based, shortest path routing algorithms.

\end{abstract}

\section{INTRODUCTION}
\label{sec:intro}
COllective INtelligences (COINs) are large, sparsely
connected recurrent neural networks, whose ``neurons'' are 
reinforcement learning (RL) algorithms. The distinguishing feature of
COINs is that their dynamics involves no centralized control, but only
the collective effects of the individual neurons each modifying their
behavior via their individual RL algorithms. This restriction holds even
though the goal of the COIN concerns the system's global behavior.
One naturally-occurring COIN is a human economy, where the ``neurons''
consist of individual humans trying to maximize their reward, and the
``goal'', for example, can be viewed as having the overall system
achieve high gross domestic product.  This paper presents a preliminary
investigation of designing and using artificial COINs as controllers
of distributed systems. The domain we consider is routing of internet traffic.

The design of a COIN starts with a global utility function specifying
the desired global behavior.  Our task is to initialize and then
update the neurons' ``local'' utility functions, without centralized
control, so that as the neurons improve their utilities, global
utility also improves. (We may also wish to update the local topology
of the COIN.) In particular, we need to ensure that
the neurons do not ``frustrate'' each other as they attempt to
increase their utilities.
The RL algorithms at each neuron that aim to optimize
that neuron's local utility are {\it microlearners}.
The learning algorithms that update the neuron's utility functions
are {\it macrolearners}.  

For robustness and breadth of applicability, we assume essentially no
knowledge concerning the dynamics of the full system, i.e., the
macrolearning and/or microlearning must ``learn'' that dynamics,
implicitly or otherwise. This rules out any approach that models the
full system.  It also means that rather than use domain knowledge to
hand-craft the local utilities as is done in multi-agent systems, in
COINs the local utility functions must be automatically initialized
and updated using only the provided global utility and (locally)
observed dynamics.

The problem of designing a COIN has never previously been addressed
in full --- hence the need for the new formalism
described below. Nonetheless, this problem is related to previous work
in many fields: distributed artificial intelligence, multi-agent
systems, computational ecologies, adaptive control, game theory~\cite{futi91},
computational markets ~\cite{baum98}, Markov decision theory, and ant-based
optimization.

For the particular problem of routing, examples of relevant work include
~\cite{boli94,chye96,kola97,mami98,sudr97}.  Most of that previous work uses
microlearning to set the internal parameters of routers running
conventional shortest path algorithms (SPAs). However the
microlearning occurs, they do not address the problem of ensuring that 
the associated local utilities do not cause the microlearners to work at cross purposes.

This paper concentrates on COIN-based setting of local utilities
rather than macrolearning. We used simulations to compare three
algorithms. The first two are an SPA and a COIN. Both had ``full
knowledge'' (FK) of the true reward-maximizing path, with reward being
the routing time of the associated router's packets for the SPAs, but
set by COIN theory for the COINs. The third algorithm was a COIN using
a memory-based (MB) microlearner~\cite{atmo96} whose knowledge was
limited to local observations.

The performance of the FK COIN was the theoretical optimum. The
performance of the FK SPA was $12.5 \pm 3 \; \%$ worse than optimum.
Despite limited knowledge, the MB COIN outperformed the FK SPA,
achieving performance $36 \pm 8 \; \%$ closer to optimum.  Note that
the performance of the FK SPA is an upper bound on the performance of
any RL-based SPA.  Accordingly, the performance of the MB COIN is at
least $36\%$ superior to that of any RL-based SPA.

Section~\ref{sec:math} below presents a cursory overview
of the mathematics behind COINs.  Section~\ref{sec:route} discusses
how the network routing problem is mapped into the COIN formalism, and
introduces our experiments.  Section~\ref{sec:res} presents results of
those experiments, which establish the power of COINs in the context
of routing problems. Finally, Section~\ref{sec:disc} presents
conclusions and summarizes future research directions.

\section{MATHEMATICS OF COINS} \label{sec:math}

The mathematical framework for COINs is quite
extensive~\cite{wotu99,wowh99a}.  This paper concentrates on four of
the concepts from that framework: subworlds, factored systems,
constraint-alignment, and the wonderful-life utility function.

We consider the state of the system across a set of discrete,
time steps, $t \in \{0, 1, ...\}$.  All 
characteristics of a neuron at time $t$ --- including its internal
parameters at that time as well as its externally visible actions ---
are encapsulated in a real-valued vector $\underline{\zeta}_{\eta,t}$. We
call this the ``state'' of neuron $\eta$ at time $t$, and let
$\underline{\zeta}$ be the state of all neurons across all time.
World utility, $G(\underline{\zeta})$, is a function of the state of
all neurons across all time, potentially not expressible as a discounted sum.

A subworld is a set of neurons.  All neurons in the same subworld
$\omega$ share the same {\it subworld utility function}
$g_{\omega}(\underline{\zeta})$. So when each subworld is a set of
neurons that have the most effect on each other, neurons are unlikely
to work at cross-purposes --- all neurons that affect each other
substantially share the same local utility.

Associated with subworlds is the concept of a (perfectly) {\it
constraint-aligned} system. In such systems any change to the neurons
in subworld $\omega$ at time 0 will have no effects on the neurons
outside of $\omega$ at times later than 0.  Intuitively, a system is
constraint-aligned if the neurons in separate subworlds do not affect
each other directly, so that the rationale behind the use of subworlds
holds.

A {\it subworld-factored} system is one where for each subworld
$\omega$ considered by itself, a change at time 0 to the states of the
neurons in that subworld results in an increased value for
$g_{\omega}(\underline{\zeta})$ if and only if it results in an
increased value for $G(\underline{\zeta})$.  For a subworld-factored
system, the side effects on the rest of the system of $\omega$'s
increasing its own utility (which perhaps decrease other
subworlds' utilities) do not end up decreasing world utility. For
these systems, the separate subworlds successfully pursuing their
separate goals do not frustrate each other as far as world utility is
concerned.

The desideratum of subworld-factored is carefully crafted. In
particular, it does {\it not} concern changes in the value of the
utility of subworlds other than the one changing its actions. Nor
does it concern changes to the states of neurons in more than one
subworld at once.  Indeed, consider the following alternative
desideratum: any change to the $t= 0$ state of the entire system that
improves all subworld utilities simultaneously also improves world
utility. Reasonable as it may appear, one can construct examples of systems
that obey this desideratum and yet quickly evolve to a {\it minimum} of
world utility~\cite{wowh99a}.

It can be proven that for a subworld-factored system, 
when each of the neurons' reinforcement learning algorithms are performing 
as well as they can, given each others' behavior, world utility
is at a critical point. 
Correct global behavior corresponds to learners reaching a (Nash)
equilibrium~\cite{kola97,wowh99b}.  There can be no tragedy of the 
commons for a subworld-factored system~\cite{hard68,wotu99,wowh99a}.

Let $\mbox{CL}_{\omega}(\underline{\zeta})$ be defined as the vector
$\underline{\zeta}$ modified by clamping the states of all neurons in
subworld $\omega$, across all time, to an arbitrary fixed value, here
taken to be 0. The {\it wonderful life
subworld utility} (WLU) is:
\begin{equation}
g_{\omega}(\underline{\zeta}) \equiv
G(\underline{\zeta}) - G(\mbox{CL}_{\omega}(\underline{\zeta}))
\end{equation}

When the system is constraint-aligned, so that, loosely speaking,
subworld $\omega$'s ``absence'' would not affect the rest of the
system, we can view the WLU as analogous to the change in world
utility that would have arisen if subworld $\omega$ ``had never
existed''.  (Hence the name of this utility - cf. the Frank Capra
movie.)  Note however, that $\mbox{CL}$ is a purely mathematical
operation.  Indeed, no assumption is even being made that
$\mbox{CL}_{\omega}(\underline{\zeta})$ is consistent with the
dynamics of the system. The sequence of states the neurons in
$\omega$ are clamped to in the definition of the WLU need not
be consistent with the dynamical laws of the system.

This dynamics-independence is a crucial strength of the WLU. 
It means that to evaluate the WLU we do {\it not} try to
infer how the system would have evolved if all neurons in $\omega$ were
set to 0 at time 0 and the system evolved from there. So long as we
know $\underline{\zeta}$ extending over all time, and so long as we
know $G$, we know the value of WLU. This is true even if we
know nothing of the dynamics of the system.

In addition to assuring the correct equilibrium behavior, there exist
many other theoretical advantages to having a system be
subworld-factored. In particular, the experiments in this paper
revolve around the following fact: a constraint-aligned system with
wonderful life subworld utilities is subworld-factored.  Combining
this with our
previous result that subworld-factored systems are at Nash equilibrium at
critical points of world utility, this result leads us to expect that
a constraint-aligned system using WL utilities in the microlearning
will approach near-optimal values of the world utility.  No such
assurances accrue to WL utilities if the system is not
constraint-aligned however. Accordingly our experiments constitute an
investigation of how well a particular system performs when WL
utilities are used but little attention is paid to ensuring that the
system is constraint-aligned.

\section{COINS FOR NETWORK ROUTING}
\label{sec:route}
\begin{figure}[thb]
\centering
\vspace*{-.1in}
\subfigure[Network A]{\label{fig:net1}
\includegraphics[width=2.3in,height=2.0in]{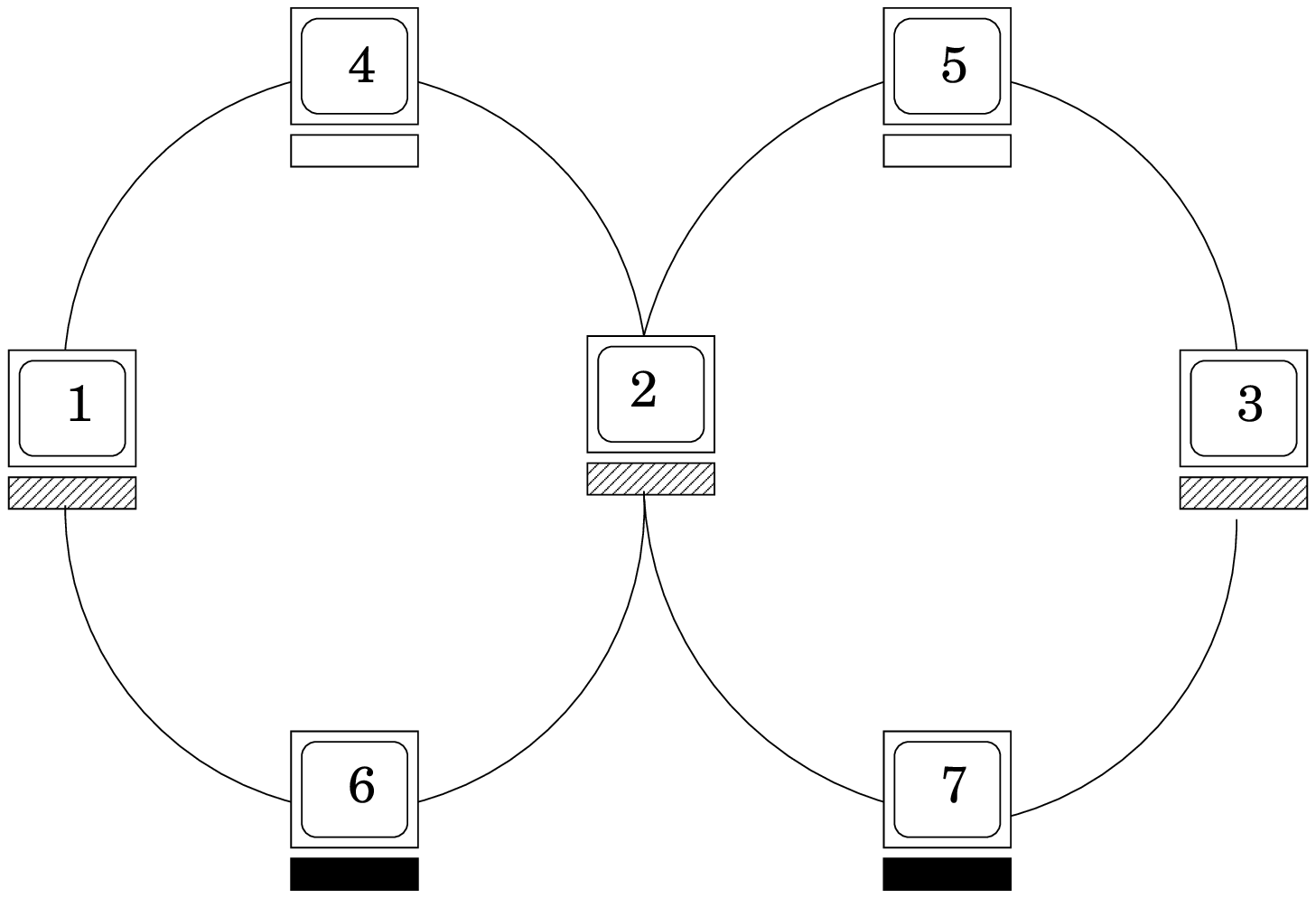} }
\subfigure[Network B]{\label{fig:net6}
\includegraphics[width=2.3in,height=2.0in]{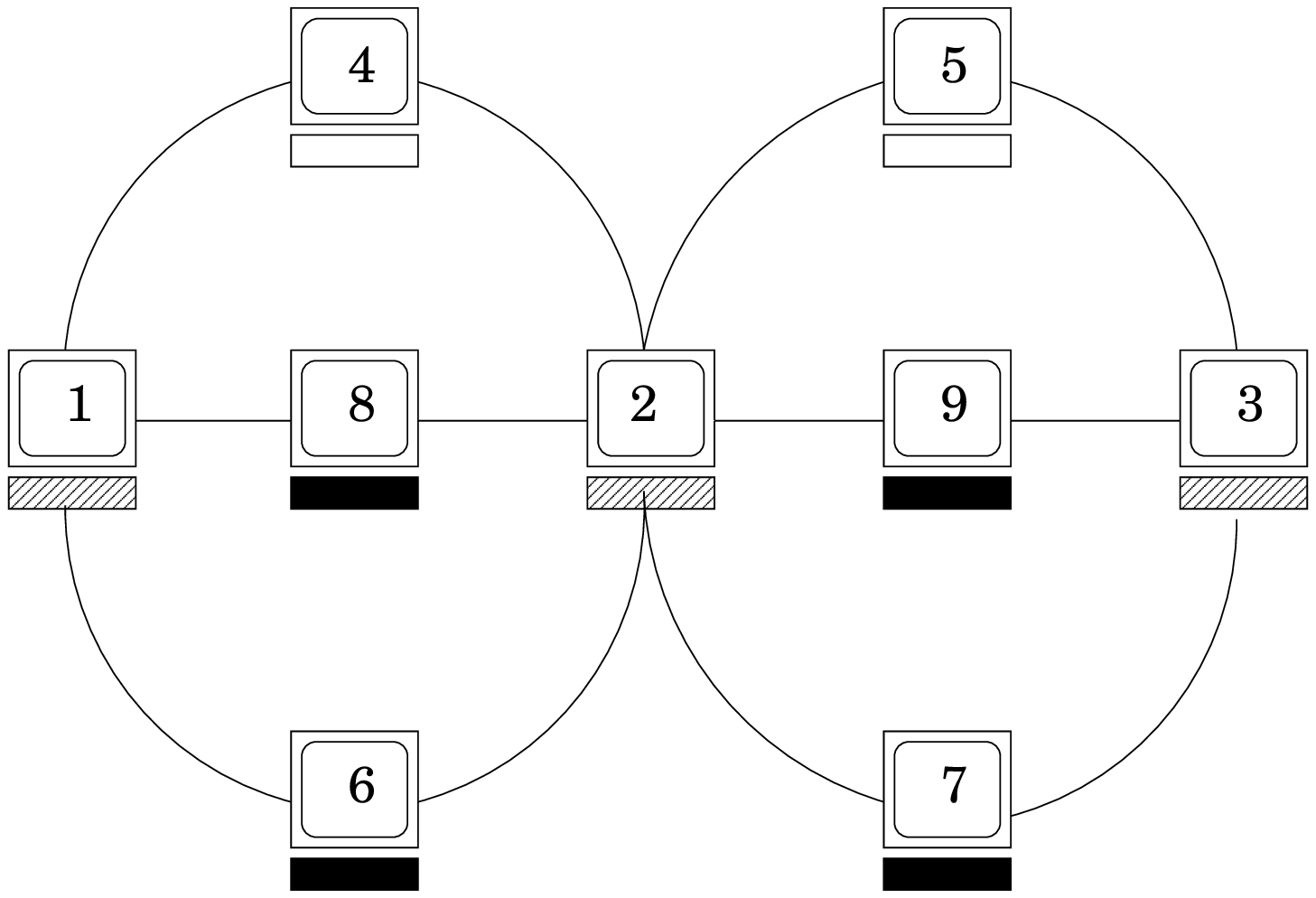} }
\vspace*{-.1in}
\caption{Network Architectures.}
\label{fig:net}
\end{figure}

In our experiments we concentrated on the two networks in 
Figure~\ref{fig:net}, both slightly larger than those in~\cite{mami98}.
To facilitate the analysis, traffic originated
only at routers indicated with white boxes and had only the routers 
indicated by dark boxes as ultimate destinations.
Note that in both networks there is a bottleneck at router 2.

As is standard in much of traffic network analysis~\cite{bega92}, at
any time all {\it traffic} at a router is a real-valued number
together with an ultimate destination tag. At each timestep, each
router sums all traffic received from upstream routers in this
timestep, to get a {\it load}. The router then decides which
downstream router to send its load to, and the cycle repeats.

A running average is kept of the total value of each router's load
over a window of the previous $L$ timesteps. This average is run
through a {\it load-to-delay} function, $W(x)$, to get the summed 
{\em delay} accrued at this timestep by all those packets traversing 
this router at this timestep.  Different routers had different
$W(x)$, to reflect the fact that real networks have
differences in router software and hardware (response time, queue
length, processing speed etc).  In our experiments $W(x) = x^3$ for
routers 1 and 3, and $W(x) = log(x+1)$ for router 2, for both
networks. The global goal is to minimize total delay encountered by
all traffic.

In terms of the COIN formalism, we identified the neurons $\eta$ as
individual pairs of routers and ultimate destinations. So
$\underline{\zeta}_{\eta,t}$ was the vector of traffic sent along all
links exiting $\eta$'s router, tagged for $\eta$'s ultimate
destination, at time $t$.  Each subworld consisted of the set all
neurons that shared a particular ultimate destination.

In the SPA each node $\eta$ tries to set $\underline{\zeta}_{\eta,t}$ 
to minimize the sum of the delays to be accrued by that traffic 
on the way to its ultimate destination.
In contrast, in a COIN $\eta$ tries to set $\underline{\zeta}_{\eta,t}$ 
to optimize $g_\omega$ for the subworld
$\omega$ containing $\eta$. For both algorithms, ``full knowledge''
means that at time $t$ all of the routers know the window-averaged
loads for all routers for time $t - 1$, and assume that those values
will be the same at $t$.  For large enough $L$, this assumption will
be arbitrarily good, and therefore will allow the
routers to make arbitrarily accurate estimates of how best to route
their traffic, according to their respective routing criteria.

In contrast, having limited knowledge, the MB COIN could only {\it
predict} the WLU value resulting from each routing decision.  More
precisely, for each router-ultimate-destination pair, the associated
microlearner estimates the map from traffic on all
outgoing links (the inputs) to WLU-based reward (the outputs -- see
below).  This was done with a single-nearest-neighbor algorithm.
Next, each router could send the packets along the
path that results in outbound traffic with the best (estimated)
reward.  However to be conservative, in these experiments we instead
had the router randomly select between that path and the path selected
by the FK SPA.

The load at router $r$ at time $t$ is determined by
$\underline{\zeta}$. Accordingly, we can encapsulate the load-to-delay
functions at the nodes by writing the delay at node $r$ at time $t$ as
$W_{r,t}(\underline{\zeta})$. In our experiments world utility was the
total delay, i.e., $G(\underline{\zeta}) = \sum_{r,t}
W_{r,t}(\underline{\zeta})$.  So using the WLU,
$g_\omega(\underline{\zeta}) = \sum_{r,t}
\Delta_{\omega,r,t}(\underline{\zeta})$, where
$\Delta_{\omega,r,t}(\underline{\zeta}) = [W_{r,t}(\underline{\zeta})
- W_{r,t}(\mbox{CL}_{\omega}(\underline{\zeta}))]$.  At each time $t$,
the MB COIN used $\sum_{r} \Delta_{\omega,r,t}(\underline{\zeta})$ as
the ``WLU-based'' reward signal for trying optimize this full WLU.

In the MB COIN, evaluating this reward in a decentralized fashion was
straight-forward. All packets have a header containing a running sum
of the $\Delta$'s encountered in all the routers it has traversed so
far. Each ultimate destination sums all such headers it received
and echoes that sum back to all routers that had routed to it.  In
this way each neuron is apprised of the WLU-based reward of its
subworld.

\section{EXPERIMENTAL RESULTS}
\label{sec:res}
The networks discussed above were tested under light, medium and heavy traffic
loads. Table~\ref{tab:traffic} shows the associated
destinations (cf. fig. 1). 

\begin{table} [htb] \centering
\vspace*{-.15in}
\caption{Source--Destination Pairings for the Three Traffic Loads}
\vspace*{.05in}
\begin{tabular}{c|c||c|c|c} \hline 
Network &  Source & Dest. (Light) & Dest. (Medium) & Dest. (Heavy) \\ \hline \hline 
 A & 4 & 6   & 6,7   & 6,7     \\
   & 5 & 7   & 7     & 6,7     \\ \hline
 B & 4 & 7,8 & 7,8,9 & 6,7,8,9 \\
   & 5 & 6,9 & 6,7,9 & 6,7,8,9 \\ \hline
\end{tabular}
\label{tab:traffic}
\end{table}

In our experiments one new packet was fed to each source router at
each time step.  Table~\ref{tab:results} reports the average total
delay (i.e., average per packet time to traverse the total network) in
each of the traffic regimes, for the shortest path algorithm with full
knowledge, the COIN with full knowledge, and the MB COIN.  Each table
entry is based on $50$ runs with a window size of $50$, and the errors
reported are errors in the mean\footnote{The results are
qualitatively identical for window sizes 20 and 100 along with total
timesteps of 100 and 500.}. All the entries in Table~\ref{tab:results}
are statistically different at the $.05$ level, including FK SPA vs. MB
COIN for Network A under light traffic conditions.

\begin{table} [htb] \centering
\vspace*{-.1in}
\caption{Average Total Delay}
\vspace*{.05in}
\begin{tabular}{c||c|c|c|c} \hline 
Network &  Load & FK SPA & FK COIN & MB COIN \\ \hline \hline 
   & light & 0.53 $\pm$ .007  & 0.45 $\pm$ .001   & 0.50 $\pm$ .008 \\
 A & medium& 1.26 $\pm$ .010  & 1.10 $\pm$ .001   & 1.21 $\pm$ .009 \\
   & heavy & 2.17 $\pm$ .012  & 1.93 $\pm$ .001   & 2.06 $\pm$ .010 \\ \hline
   & light & 2.13 $\pm$ .012  & 1.92 $\pm$ .001   & 2.05 $\pm$ .010 \\
 B & medium& 4.37 $\pm$ .014  & 3.96 $\pm$ .001   & 4.19 $\pm$ .012 \\
   & heavy & 6.94 $\pm$ .015  & 6.35 $\pm$ .001   & 6.82 $\pm$ .024 \\ \hline
\end{tabular}
\label{tab:results}
\end{table}

Table~\ref{tab:results} provides two important observations: First,
the WLU-based COIN outperformed the SPA when both have full knowledge,
thereby demonstrating the superiority of the new routing strategy. By
not having its routers greedily strive for the shortest paths for
their packets, the COIN settles into a more desirable state that
reduces the average total delay for {\em all} packets.  Second, even
when the WLU is estimated through a memory-based learner (using only
information available to the local routers), the performance of the
COIN still surpasses that of the FK SPA.  This result not only
establishes the feasibility of COIN-based routers, but also
demonstrates that for this task COINs will outperform {\it any}
algorithm that can only estimate the shortest path, since the
performance of the FK SPA is a ceiling on the performance of any
such RL-based SPA.

\begin{figure}[bth]
\centering
\vspace*{-.15in}
\subfigure[Network A]{\label{fig:res1}
\includegraphics[width=2.4in,height=2.0in]{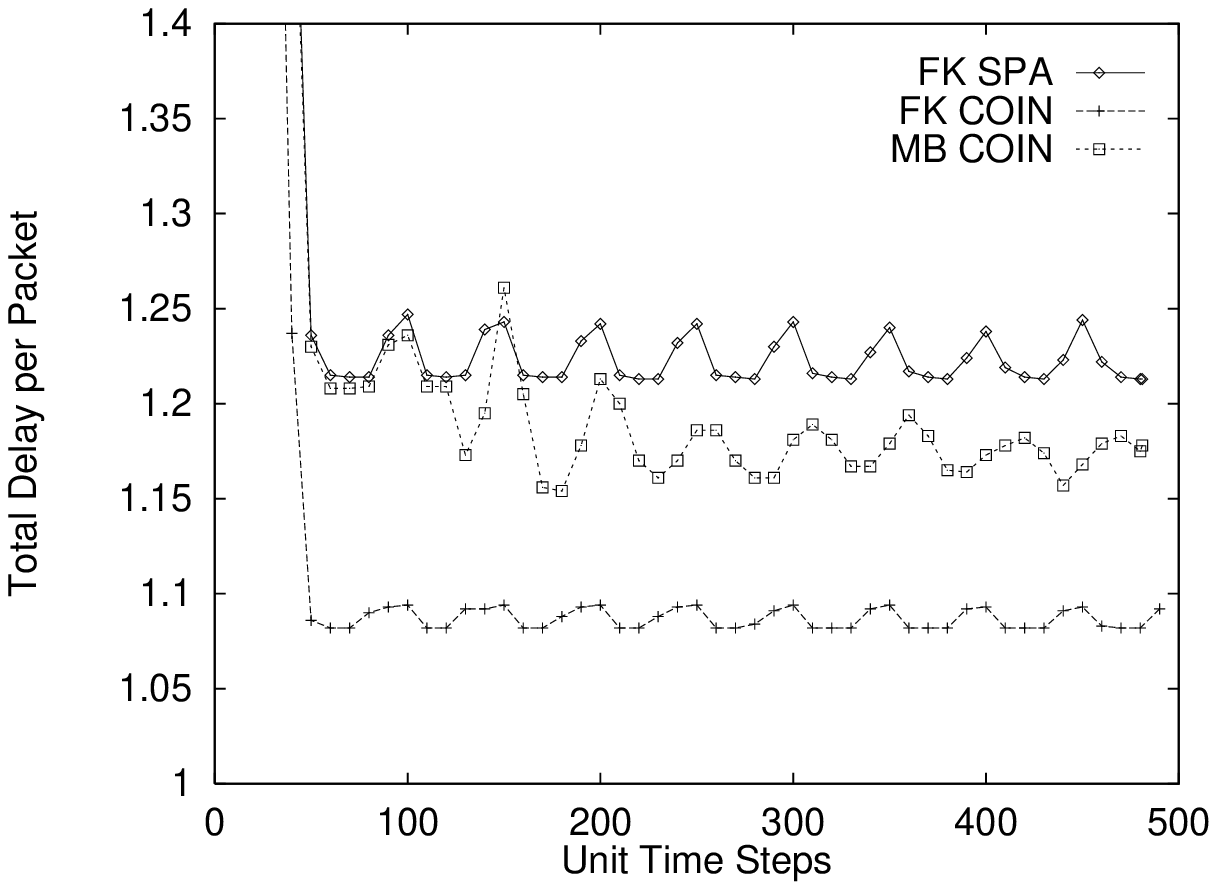}}
\subfigure[Network B]{\label{fig:res6}
\includegraphics[width=2.4in,height=2.0in]{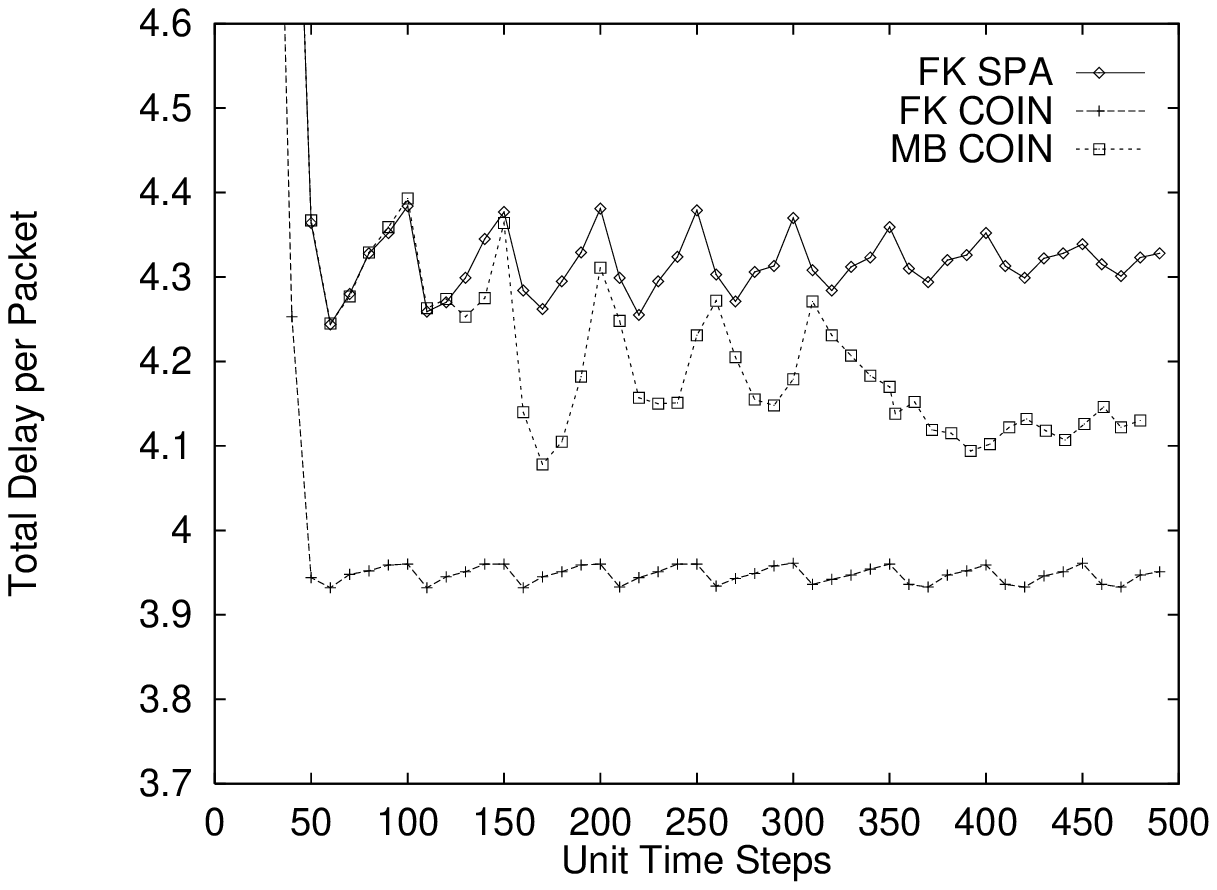}}
\vspace*{-.2in}
\caption{Total Delay.}
\vspace*{-.25in}
\label{fig:res}
\end{figure}

Figure~\ref{fig:res} shows how total delay varies with time for the
medium traffic regime (each plot is based on $50$ runs). The
``ringing'' is an artifact caused by the starting conditions and the
window size ($50$).  Note that for both networks the FK COIN not only
provides the shortest delays, but also settles into that solution very
rapidly. 

\section{DISCUSSION}
\label{sec:disc}
Many distributed computational tasks are naturally addressed as
recurrent neural networks of reinforcement learning algorithms (i.e.,
COINs). The difficulty in doing so is ensuring that, despite the
absence of centralized communication and control, the reward functions
of the separate neurons work in synchrony to foster good global performance,
rather than cause their associated neurons to work at cross-purposes.

The mathematical framework synopsized in this paper is a theoretical
solution to this difficulty. To assess its real-world applicability,
we employed it to design a full-knowledge (FK) COIN as well as a
memory-based (RL-based) COIN, for the task of packet routing on a
network. We compared the performance of those algorithms to that of a
FK shortest-path algorithm (SPA).  Not only did the FK COIN beat the
FK SPA, but also the memory-based COIN, despite having only limited
knowledge, beat the full-knowledge SPA.  This latter result is all the
more remarkable in that the performance of the FK SPA is an upper
bound on the performance of previously investigated RL-based routing
schemes, which use the RL to try to provide accurate knowledge to an
SPA.

There are many directions for future work on COINs, even restricting
attention to domain of packet routing.  Within that particular domain,
currently we are extending our experiments to larger networks, using
industrial event-driven network simulators.  Concurrently, we are
investigating the use of macrolearning for COIN-based packet-routing,
i.e., the run-time modification of the neurons' utility functions to
improve the subworld-factoredness of the COIN.

\nocite{wowh99b}

\bibliographystyle{plain}

\begin{thebibliography}{10}

\bibitem{atmo96}
C.~G. Atkenson, A.~W. Moore, and S.~Schaal.
\newblock Locally weighted learning.
\newblock {\em Artificial Intelligence Review}, Submitted, 1996.

\bibitem{baum98}
E.~Baum.
\newblock Manifesto for an evolutionary economics of intelligence.
\newblock In C.~M. Bishop, editor, {\em Neural Networks and Machine Learning}.
  Springer-Verlag, 1998.

\bibitem{bega92}
D.~Bertsekas and R.~Gallager.
\newblock {\em Data Networks}.
\newblock Prentice Hall, NJ, 1992.

\bibitem{boli94}
J.~Boyan and M.~Littman.
\newblock Packet routing in dynamically changing networks: A reinforcement
  learning approach.
\newblock In {\em Advances in
  Neural Information Processing Systems - 6}, pages 671--678. Morgan Kaufmann,
  1994.

\bibitem{chye96}
S.~P.~M. Choi and D.~Y. Yeung.
\newblock Predictive {Q}-routing: A memory based reinforcement learning
  approach to adaptive traffic control.
\newblock In {\em
  Advances in Neural Information Processing Systems - 8}, pages 945--951. MIT
  Press, 1996.

\bibitem{futi91}
D.~Fudenberg and J.~Tirole.
\newblock {\em Game Theory}.
\newblock MIT Press, Cambridge, MA, 1991.

\bibitem{hard68}
G.~Hardin.
\newblock The tragedy of the commons.
\newblock {\em Science}, 162:1243--1248, 1968.

\bibitem{kola97}
Y.~A. Korilis, A.~A. Lazar, and A.~Orda.
\newblock Achieving network optima using {S}tackelberg routing strategies.
\newblock {\em IEEE Tran. on Networking}, 5(1):161--173, 1997.

\bibitem{mami98}
P.~Marbach, O.~Mihatsch, M.~Schulte, and J.~Tsisiklis.
\newblock Reinforcement learning for call admission control and routing in
  integrated service networks.
\newblock In {\em  Adv. in Neural Info. Proc. Systems - 10},
  pages 922--928. MIT Press, 1998.

\bibitem{sudr97}
D.~Subramanian, P.~Druschel, and J.~Chen.
\newblock Ants and reinforcement learning: A case study in routing in dynamic
  networks.
\newblock In {\em Proceedings of the Fifteenth International Conference on
  Artificial Intelligence}, pages 832--838, 1997.

\bibitem{wotu99}
D.~Wolpert and K.~Tumer.
\newblock Collective {I}ntelligence.
\newblock In J.~M. Bradshaw, editor, {\em Handbook of Agent technology}. AAAI
  Press/MIT Press, 1999.
\newblock to appear.

\bibitem{wowh99a}
D.~Wolpert, K.~Wheeler, and K.~Tumer.
\newblock Automated design of multi-agent systems.
\newblock In {\em Proc. of the 3rd Int. Conf. of
  Autonomous Agents}, 1999.
\newblock to appear.

\bibitem{wowh99b}
D.~Wolpert, K.~Wheeler, and K.~Tumer.
\newblock Collective intelligence for distributed control.
\newblock 1999.
\newblock (pre-print).

\end{thebibliography}

\end{document}